\documentclass[11pt,a4paper]{article}
\usepackage[hyperref]{acl2020}
\usepackage{times}
\usepackage{latexsym}

\usepackage{graphicx}
\usepackage{amsmath}
\usepackage{booktabs}
\usepackage{multirow}
\usepackage{hyperref}
\usepackage{csquotes}

\usepackage{microtype}

\aclfinalcopy 

\title{Entity-Enriched Neural Models for Clinical Question Answering}

\author{Bhanu Pratap Singh Rawat\textsuperscript{1,}\thanks{~~The author did this work while interning at MIT-IBM Watson AI Lab.}, Wei-Hung Weng\textsuperscript{2,3,\#}, So Yeon Min\textsuperscript{2,3,$\mathsection$},\\ \textbf{Preethi Raghavan\textsuperscript{3,4,$\dagger$}, Peter Szolovits\textsuperscript{2,3,$\ddagger$}}\\
  \textsuperscript{1}UMass-Amherst, \textsuperscript{2}MIT CSAIL, \textsuperscript{3}MIT-IBM Watson AI Lab, \textsuperscript{4}IBM Research, Cambridge \\
  \texttt{\textsuperscript{*}brawat@umass.edu,
  \textsuperscript{\#}ckbjimmy@mit.edu,
  \textsuperscript{$\mathsection$}symin95@mit.edu}\\
  
  \texttt{\textsuperscript{$\dagger$}praghav@us.ibm.com, \textsuperscript{$\ddagger$}psz@mit.edu} \\} 
\date{}

\begin{document}
\maketitle

\begin{abstract}
We explore state-of-the-art neural models for question answering on electronic medical records and improve their ability to generalize better on previously unseen (paraphrased) questions at test time. We enable this by learning to predict logical forms as an auxiliary task along with the main task of answer span detection. The predicted logical forms also serve as a rationale for the answer. Further, we also incorporate medical entity information in these models via the ERNIE \cite{zhang2019ernie} architecture. We train our models on the large-scale 
emrQA dataset and observe that our multi-task entity-enriched models generalize to paraphrased questions $\sim5$\% better than the baseline BERT model. 
\end{abstract}

\section{Introduction}
\label{sec:intro}


The field of question answering (QA) has seen significant progress with several resources, models and benchmark datasets. Pre-trained neural language encoders like BERT \cite{devlin2018bert} 
and its variants \cite{seo2016bidirectional, zhang2019sg} have achieved near-human or even better performance on popular open-domain QA tasks such as SQuAD 2.0 
  \cite{rajpurkar2016squad}. 
While there has been some progress in biomedical QA on medical literature \cite{vsuster2018clicr, tsatsaronis2012bioasq}, existing models have not been similarly adapted to clinical domain on electronic medical records (EMRs).


\begin{figure}[!htbp]
\centering
\includegraphics[width=8cm]{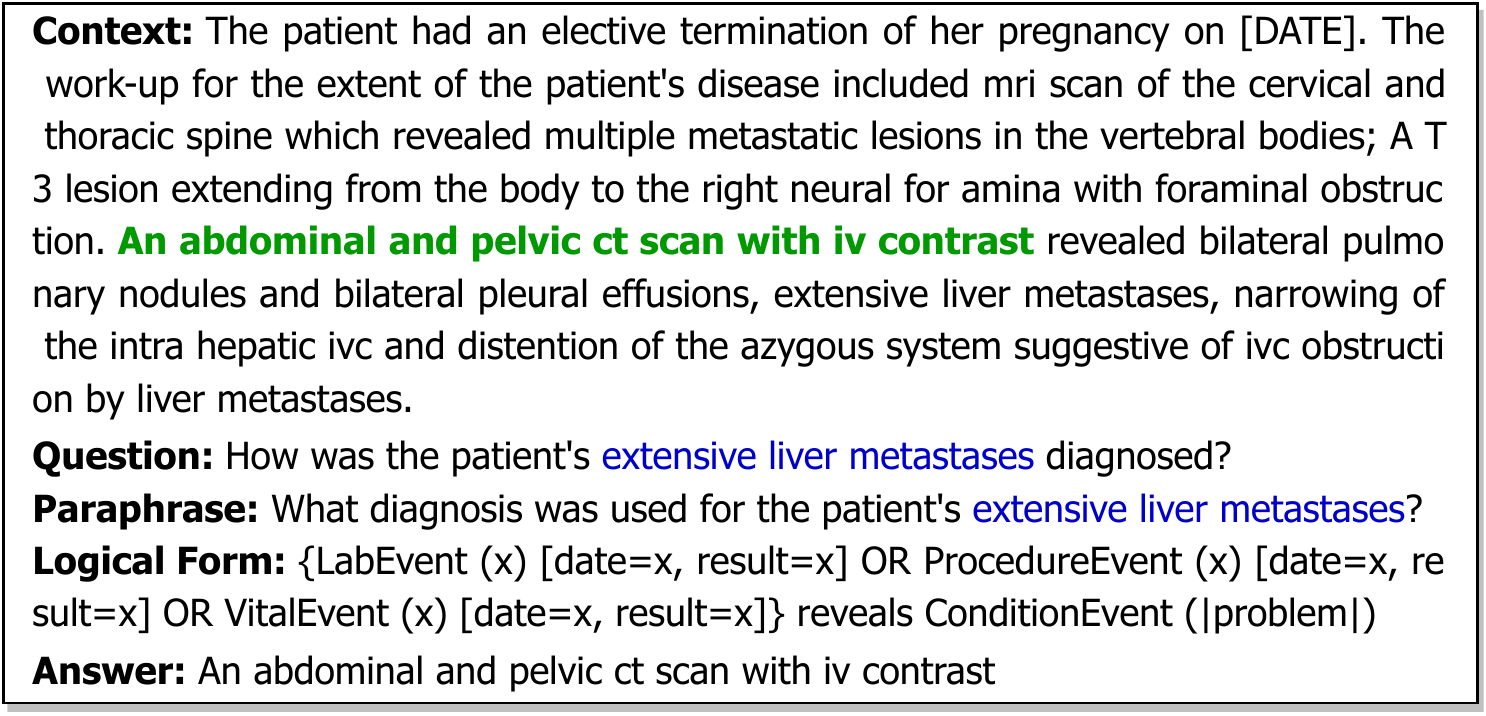}
\caption{A synthetic example of a clinical context, question, its logical form and the expected answer.}
\label{fig:emrqa_qa}
\end{figure}

 Community-shared large-scale datasets like emrQA \cite{pampari2018emrqa} allow us to apply state-of-the-art models, establish benchmarks, innovate and adapt them to clinical domain-specific needs. emrQA enables question answering from electronic medical records (EMRs) where a question is asked by a physician against a patient's medical record (clinical notes). Thus, we adapt these models for EMR QA while focusing on model generalization via the following.  (1) learning to predict the logical form (a structured semantic representation that captures the answering needs corresponding to a natural language question) along with the answer and (2) incorporating medical entity embeddings into models for EMR QA. We now examine the motivation behind these.



A physician interacting with a QA system on EMRs may ask the same question in several different ways; 
a physician may frame a question as:
\textit{``Is the patient allergic to penicillin?''} whereas the other could frame it as \textit{``Does penicillin cause any allergic reactions to the patient?''}. Since paraphrasing is a common form of generalization in natural language processing (NLP) \cite{bhagat2009acquiring}, a QA model should be able to generalize well to such paraphrased question variants that may not be seen during training (and avoid simply memorizing the questions). However, current state-of-the-art models do not consider the use of meta-information such as the semantic parse or logical form of the questions in unstructured QA. In order to give the model the ability to understand the semantic information about answering needs of a question, we frame our problem in a multitask learning setting where the primary task is extractive QA and the auxiliary task is the logical form prediction of the question.
Fine-tuning on medical copora (MIMIC-III, PubMed \cite{johnson2016mimic, lee2020biobert}) helps models like BERT align their representations according to medical vocabulary (since they are previously trained on open-domain corpora such as WikiText 
\cite{zhu2015aligning}). However, another challenge for developing EMR QA models is that different physicians can use different medical terminology to express the same entity; e.g., ``heart attack'' vs.  ``myocardial infarction''. 
Mapping these phrases to the same UMLS semantic type\footnote{\url{https://metamap.nlm.nih.gov/SemanticTypesAndGroups.shtml}} as \textit{disease or syndrome (dsyn)} provides common information between such medical terminologies. Incorporating such entity information about tokens in the context and question can further improve the performance of QA models for the clinical domain.

Our contributions are as follows: 
\vspace{-0.3em}
\begin{enumerate}
    \item We establish state-of-the-art benchmarks for EMR QA on a large clinical question answering dataset, emrQA \cite{pampari2018emrqa} 
    
    \item We demonstrate that incorporating an auxiliary task of predicting the logical form of a question helps 
    the proposed models generalize well over unseen paraphrases, improving the overall performance on emrQA by $\sim5\%$ over BERT \cite{devlin2018bert} and by $\sim3.5\%$ over clinicalBERT \cite{alsentzer2019publicly}. We support this hypothesis by running our proposed model over both emrQA and another clinical QA dataset, MADE \cite{jagannatha2019overview}.
    
    \item The predicted logical form for unseen paraphrases helps in understanding the model better and provides a rationale (explanation) for why the answer was predicted for the provided question. This information is critical in \emph{clinical domain} as it provides an accompanying answer justification for  clinicians.
    
    \item We incorporate medical entity information by including entity embeddings via the ERNIE \cite{zhang2019ernie}  architecture \cite{zhang2019ernie} and observe that the model accuracy and ability to generalize goes up by $\sim3\%$ over $\text{BERT}_{base}$\cite{devlin2018bert}. 
\end{enumerate}

\section{Problem Formulation}
\label{sec:problem_formulation}
We formulate the EMR QA problem as a reading comprehension task. Given a natural language  \emph{question} (asked by a physician) and a \emph{context}, where the context is a set of contiguous sentences from a patient's EMR (unstructured clinical notes), the task is to predict the answer span from the given context. 
Along with the $($question, context, answer$)$ triplet, also available as input are \emph{clinical entities} extracted from the question and context.
Also available as input is the, \emph{logical form} (LF) 
that is a structured representation that captures answering needs of the question through entities, attributes and relations required to be in the answer \cite{pampari2018emrqa}. A question may have multiple paraphrases where all paraphrases map to the same LF (and the same answer, fig.~\ref{fig:emrqa_qa}). 


\begin{figure*}[!htbp]
\centering
\includegraphics[width=14cm]{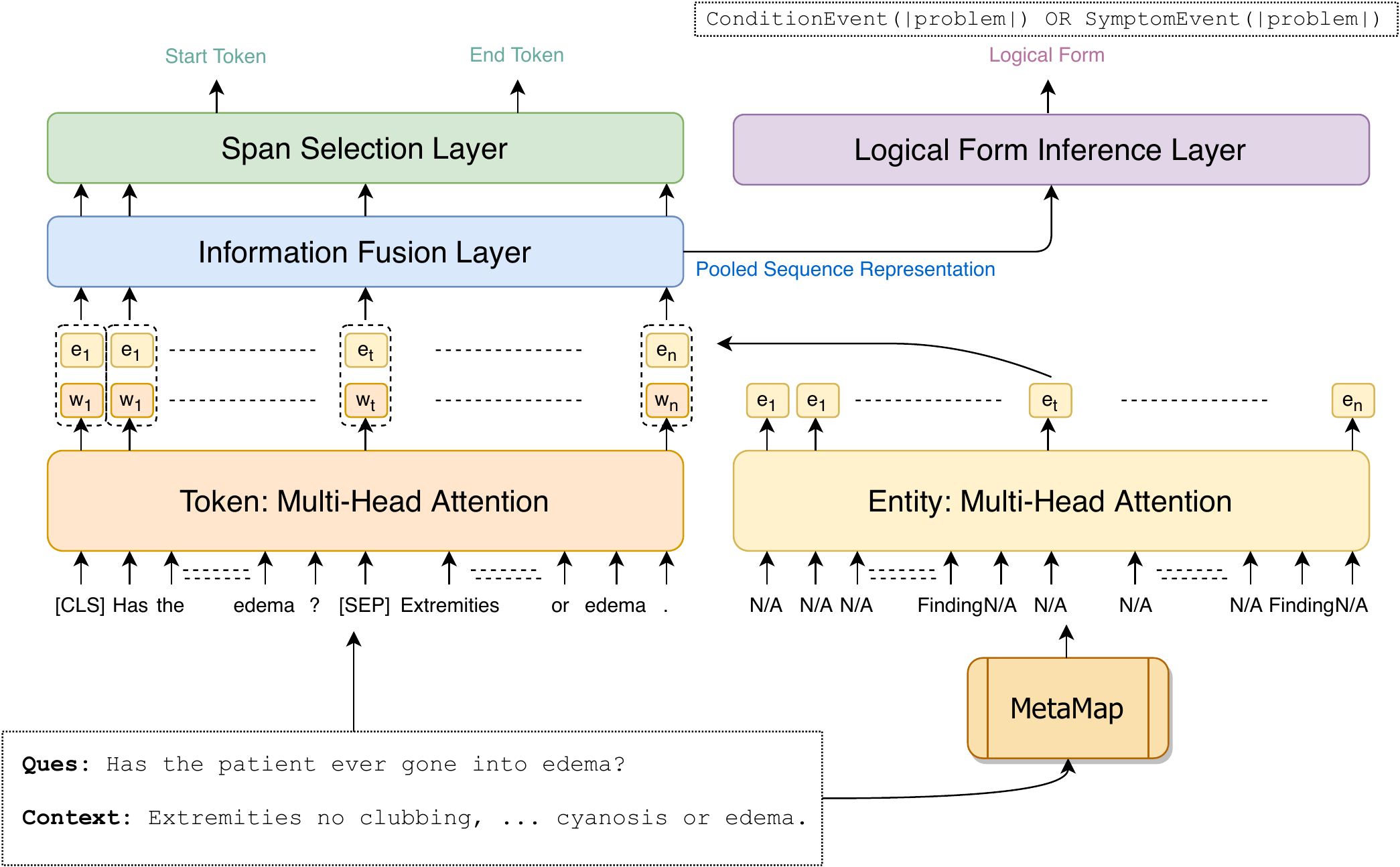}
\caption{The network architecture of our multi-task learning question answering model (\emph{M-cERNIE}). The question and context are provided to a multi-head attention model (\emph{orange}) and are also passed through MetaMap to extract clinical entities which are passed through a separate multi-head attention (\emph{yellow}). The token and entity representations are then passed through an information fusion layer (\emph{blue}) to extract entity-enriched token representations which are then used for answer span prediction. The pooled sequence representation from the information fusion layer is passed through logical form inference layer to predict the logical form.}
\label{fig:models_png}
\end{figure*}

\section{Methodology}
\label{sec:methodology}
In this section, we briefly describe BERT \cite{devlin2018bert}, ERNIE \cite{zhang2019ernie} and our proposed model.

\subsection{Bidirectional Encoder Representations from Transformers (BERT)}
BERT \cite{devlin2018bert} uses multi-layer bidirectional Transformer \cite{vaswani2017attention} networks to encode contextualised language representations. BERT representations are learned from two tasks: masked language modeling \cite{taylor1953cloze} and next sentence prediction task. We chose BERT model as pre-trained BERT models can be fine-tuned with just one additional inference layer and it achieved state-of-the-art results for a wide range of tasks such as question answering, such as SQuAD \cite{rajpurkar2016squad, rajpurkar2018know}, and multiple language inference tasks, such as MultiNLI \cite{williams2017broad}. 
\emph{clinicalBERT} \cite{alsentzer2019publicly} 
yielded superior performance on clinical-related NLP tasks such as i2b2 named entity recognition (NER) challenges \cite{uzuner20112010}. It was created by further fine-tuning of $\text{BERT}_{base}$ with biomedical and clinical corpus (MIMIC-III) \cite{johnson2016mimic}. 


\subsection{Enhanced Language Representation with Informative Entities (ERNIE)}
We adopt the ERNIE framework \cite{zhang2019ernie} to integrate the entity-level clinical concept information into the BERT architecture, which has not yet been explored in the previous works. ERNIE has shown significant improvement in different entity typing and relation classification tasks, as it utilises the extra entity information which is provided from knowledge graphs. ERNIE uses BERT for extracting contextualized token embeddings and a multi-head attention model to generate entity embeddings. These two set of embeddings are aligned and provided as an input to an information fusion layer which provides entity-enriched token embeddings. For a token ($w_j$) and its aligned entity ($e_k = f(w_j)$), the information fusion process is as follows:
\begin{equation}
    \label{eq:ernie_inf_fusion}
    h_j = \sigma (W_t^{(i)}w_j^{(i)} + W_e^{(i)}e_k^{(i)} + b^{(i)})
\end{equation}
Here $h_j$ represents the entity enriched token embedding, $\sigma$ is the non-linear activation function, $W_t$ refers to an affine layer for token embeddings and $W_e$ refers to an affine layer for entity embeddings. For the tokens without corresponding entities, the information fusion process becomes:
\begin{equation}
    h_j = \sigma (W_t^{(i)}w_j^{(i)} + b^{(i)})
\end{equation}
Initially, each entity embedding is assigned randomly and is fine-tuned along with token embeddings throughout the training procedure. The ERNIE architecture would be applicable to the model even if the logical forms are not available.




\subsection{Multi-task Learning for Extractive QA}
In order to improve the ability of a QA model to generalize better over paraphrases, it helps to provide the model information about the logical form that links these paraphrases. Since the answer to all the paraphrased questions is the same (and hence, logical form is the same), we constructed a multi-task learning framework to incorporate the logical form information into the model. 
Thus, along with predicting the answer span, we added an auxiliary task to also predict the corresponding logical form of the question. Multi-task learning provides an inductive bias to enhance the primary task's performance via auxiliary tasks \cite{weng2019multimodal}. In our setting, the primary task is span detection of the answer and the auxiliary task is logical form prediction for both emrQA and MADE (both datasets are explained in detail in \S~\ref{sec:dataset}). The final loss for our model is defined as:
\begin{equation}
    \mathcal{L}_{model} = \omega \mathcal{L}_{lf} + (1-\omega) \mathcal{L}_{span},
\end{equation}
where $\omega$ is the weightage given to the loss of auxillary task ($\mathcal{L}_{lf}$), logical form prediction. $\mathcal{L}_{span}$ is loss for answer span prediction and $\mathcal{L}_{model}$ is the final loss for our proposed model. The multi-task learning model can work with both BERT and ERNIE as the base model. Figure~\ref{fig:models_png} depicts the proposed multi-task model to predict both the answer and logical form given a question and ERNIE architecture that is used to learn entity-enriched token embeddings.


\section{Datasets}
\label{sec:dataset}
We used \emph{emrQA}\footnote{\url{https://github.com/panushri25/emrQA}} and \emph{MADE}\footnote{\url{https://bio-nlp.org/index.php/projects/39-nlp-challenges}} datasets for our experiments. 
We provide a brief summary of each dataset and the methodology followed to split these datasets into train and test sets. 

\paragraph{emrQA} 
The emrQA corpus \cite{pampari2018emrqa} is the only community-shared clinical QA dataset that consists of questions, posed by physicians against electronic medical records (EMRs) of a patient, along with their answers. The dataset was developed by leveraging existing annotations available for other clinical natural language processing (NLP) tasks (i2b2 challenge datasets \cite{uzuner20112010}). It is a credible resource for clinical QA as logical forms that are generated by a physician help slot fill question templates and extract corresponding answers from annotated notes. Multiple question templates can be mapped to the same logical form (LF), as shown in Table~\ref{tab:paraphrases}, and are referred to as \emph{paraphrases} of each other.

\begin{table}[!htbp]
\footnotesize
\begin{tabular}{l}
\toprule
\textbf{LF:} MedicationEvent ($|$medication$|$) [dosage=x] 
\\\midrule 
How much $|$medication$|$ does the patient take per day? \\
 What is her current dose of $|$medication$|$?\\
 What is the current dose of the patient's $|$medication$|$?\\
 What is the current dose of $|$medication$|$?\\
 What is the dosage of $|$medication$|$?\\
 What was the dosage prescribed of $|$medication$|$?\\ \bottomrule
\end{tabular}
\caption{A logical form (LF) and its respective question templates (paraphrases).}
\label{tab:paraphrases}
\end{table}

The \emph{emrQA} corpus has over $1M+$ question, logical form, and answer/evidence triplets, an example of a context, question, its logical form and a paraphrase is shown in Fig~\ref{fig:emrqa_qa}. The evidences are the sentences from the clinical note that are relevant to a particular question. There are total $30$ logical forms in the emrQA dataset \footnote{\url{https://github.com/panushri25/emrQA/blob/master/templates/templates-all.csv}}. 



\paragraph{MADE}
\emph{MADE} 1.0 \cite{jagannatha2019overview} dataset was hosted as an adverse drug reactions (ADRs) and medication extraction challenge from EMRs. This dataset was converted into a QA dataset by following the same procedure as enumerated in the literature of \emph{emrQA} \cite{pampari2018emrqa}. \emph{MADE} QA dataset is smaller than \emph{emrQA}, as \emph{emrQA} consists of multiple datasets taken from i2b2 \cite{uzuner20112010} whereas \emph{MADE} only has specific relations and entity mentions to that of ADRs and medications. This resulted in a clinical QA dataset which has different properties as compared to \emph{emrQA}. \emph{MADE} also has lesser number of logical forms (8 LFs) as compared to \emph{emrQA} because of fewer entities and relations. The 8 LFs for MADE are provided in Appendix~\ref{appendix:lf_made}.

\subsection{Train/test splits}
The emrQA dataset is generated using a semi-automated process that normalizes real physician questions to create question templates, associates expert annotated logical forms with each template and slot fills them using annotations for various NLP tasks from i2b2 challenge datasets (for e.g., fig.~\ref{fig:emrqa_qa}). emrQA is rich in paraphrases as physicians often tend to express the same information need in different ways. As shown in Table.~\ref{tab:paraphrases}, all paraphrases of a question map to the same logical form.
Thus, if a model has observed some of the paraphrases it should be able to generalize to the others effectively with the help of their shared logical form ``MedicationEvent ($|$medication$|$) [dosage=x]''. In order to simulate this, and test the true capability of the model to generalize to unseen paraphrased questions, we create a splitting scheme and refer to it as \emph{paraphrase-level} split.

\paragraph{Paraphrase-level split}~\\~
The basic idea is that some of question templates would be observed by the model during training and remaining would be used during validation and testing. The steps taken for creating this split are enumerated below:
\begin{enumerate}
    \item First, the clinical notes are separated into train, val and test sets. Then the question, logical form and context triplets are generated for each set resulting in the full dataset. Here the context is the set of contiguous sentences from the EMR.
    \item Then for each logical form (LF), $70\%$ of its corresponding question templates are chosen for train dataset and the rest are kept for validation and test dataset. Considering the LF shown in Table~\ref{tab:paraphrases}, four of the question templates ($QT_{tr}$) would be assigned for training and two ($QT_{v/t}$) of them would be assigned for validation/testing. So any sample in training dataset whose question is generated from the question template set $Q_{v/t}$ would be discarded. Similarly, any sample with a question generated from the question template set $Q_{tr}$ would be discarded.
    \item To compare the generalizability performance of our model, we keep the training dataset with both set of question templates ($QT_{tr}+QT_{v/t}$) as well. Essentially, a baseline model which has observed all the question templates ($QT_{tr}+QT_{v/t}$) should be able to perform better on the $QT_{v/t}$ set as compared to a model which has only observed $QT_{tr}$ set. This comparison would help us in measuring the improvement in performance with the help of logical forms even when a set of question templates are not observed by the model.
\end{enumerate}
The dataset statistics for both \emph{emrQA} and \emph{MADE} are shown in Table~\ref{tab:data_splits}. The training set with both question template sets ($QT_{tr}+QT_{v/t}$) is shown with `(r)' appended as suffix, as it is essentially a \emph{random} split, whereas the training set with the question template ($QT_{tr}$) is appended with `(pl)' for \emph{paraphrase-level} split.

\begin{table}[!htbp]
\centering
\small
\begin{tabular}{l|lccc}
\toprule
  \textbf{Datasets} & \textbf{Split} &\textbf{Train} & \textbf{Val.} & \textbf{Test} \\
    \midrule
\multirow{2}{*}{\textbf{emrQA}}    &\# Notes &433 & 44 & 47 \\ 
&\# Samples (pl) & 133,589         & 21,666          & 19,401                 \\ 
&\# Samples (r) & 198,118         & 21,666          & 19,401                 \\ \midrule
 \multirow{2}{*}{\textbf{MADE}} &   \# Notes &788 & 88 & 213 \\ 
&\# Samples (pl) & 73,224         & 4,806          & 9,235                 \\ 
&\# Samples (r) & 113,975         & 4,806          & 9,235                 \\ \bottomrule
\end{tabular}
\caption{Train, validation and test data splits.}
\label{tab:data_splits}
\end{table}

\section{Experiments}
In this section, we briefly discuss the experimental settings, clinical entity extraction method, implementation details of our proposed model and evaluation metrics for our experiments.
\subsection{Experimental Setting}
\label{sec:problem_formulation}
As a reading comprehension style task, the model has to identify the span of the answer given the question-context pair. For both \emph{emrQA} and \emph{MADE} dataset, the span is marked as the answer to the question and the sentence is marked as the evidence. Hence, we perform extractive question answering at two levels: \textit{sentence} and \emph{paragraph}. 

\paragraph{Sentence setting:} For this setting, the evidence sentence which contains the answer span is provided as the context to the question and the model has to predict the span of the answer, given the question. 

\paragraph{Paragraph setting:}  
Clinical notes are noisy and often contain incomplete sentences, lists and embedded tables making it difficult to segment paragraphs in notes.  Hence, we decided to define the context as evidence sentence and $15-20$ sentences around it. We randomly chose the length of the paragraph ($l_{para}$) and another number less than the length of the paragraph ($l_{pre} < l_{para}$). We chose $l_{pre}$ contiguous sentences which exist prior to the evidence sentence in the EMR and ($l_{para} - l_{pre}$) sentences after the evidence sentence. We adopted this strategy because the model could have benefited from the information that the evidence sentence is exactly in the middle of a fixed length paragraph. The model has to predict the span of the answer from the $l_{para}$ sentences long paragraph (context) given the question.

The datasets are appended by `-p' and `-s' for \emph{paragraph} and \emph{sentence} settings respectively. The \emph{sentence} setting is a relatively easier setting, for the model, compared to the \emph{paragraph} setting because the scope of the answer is narrowed down to lesser number of tokens and there is less noise. For both settings, as also mentioned in \S~\ref{sec:dataset}, we kept the train set where all the question templates (paraphrases) are observed by the model during training and that is referred with `(r)' prefix, suggesting `random' selection and no filtering based on question templates (paraphrases). All these dataset abbreviations are shown in the first column of Table~\ref{tab:exact_match_results}.

\subsection{Extracting Entity Information}
\label{sec:metamap}
\verb+MetaMap+ \cite{aronson2001effective} uses a knowledge-intensive approach to discover different clinical concepts referred to in the text according to unified medical language system (UMLS) \cite{bodenreider2004unified}. The clinical ontologies, such as SNOMED \cite{spackman1997snomed} and RxNorm \cite{liu2005rxnorm}, embedded  in \verb+MetaMap+ are quite useful in extracting $\sim127$ entities across diagnosis, medication, procedure and sign/symptoms. We shortlisted these entities (semantic types) by mapping them to the entities which were used for creating logical forms of the questions as these are the main entities for which the question has been posed. The selected entities are: acab, aggp, anab, anst, bpoc, cgab, clnd, diap, emod, evnt, fndg, inpo, lbpr, lbtr, phob, qnco, sbst, sosy and topp. Their descriptions are provided in Appendix~\ref{appendix:sem_types}.

These filtered entities (Table~\ref{tab:sem_defs}), extracted from \verb+MetaMap+, are provided to ERNIE. A separate embedding space is defined for the entity embeddings which are passed through a multi-head attention layer \cite{vaswani2017attention} before interacting with token embeddings in the information fusion layer. The entity-enriched token embeddings are then used to predict the span of the answer from the context. We fine-tuned these entity embeddings along with the token embeddings, as opposed to using learned entities and not fine-tuning during downstream tasks \cite{zhang2019ernie}. The architecture is illustrated in Fig~\ref{fig:models_png}.

\subsection{Implementation Details}
The BERT model was released with pre-trained weights as $\text{BERT}_{base}$ and $\text{BERT}_{large}$. $\text{BERT}_{base}$ has lesser number of parameters but achieved state-of-the-art results on a number of open-domain NLP tasks. We performed our experiments with $\text{BERT}_{base}$ and hence, from here onwards we refer to $\text{BERT}_{base}$ as \emph{BERT}. A fine-tuned version of $\text{BERT}_{base}$ on clinical notes was released as clinicalBERT (\emph{cBERT}) \cite{alsentzer2019publicly}. We use \emph{cBERT} as the multi-head attention model for getting the token representations in ERNIE. We refer to this version of ERNIE, with entities from \verb+MetaMap+, as \emph{cERNIE} for clinical ERNIE. Our final multi-task learning model, incorporated with an auxillary task of predicting logical forms, is referred to as \emph{M-cERNIE} for multi-task clinical ERNIE. The code for all the models is provided at  \url{https://github.com/emrQA/bionlp_acl20}. 

\paragraph{Evaluation Metrics} For our extractive question answering task, we utilised exact match and F1-score for evaluation as per earlier literature \cite{rajpurkar2016squad}.

\section{Results and Discussion}
\label{sec:results_discussion}
In this section, we compare the results of all the models that we introduced in \S~\ref{sec:methodology}. With the help of different experiments, we try to analyse whether the induced entity and logical form information help the model in achieving better performance or not. We also analyse the logical form predictions to understand whether it provides a rationale for the answer predicted by our proposed model. The compiled results for all the models are shown in Table~\ref{tab:exact_match_results}. The hyper-parameter values for the best performing models are provided in Appendix~\ref{appendix:hyper}.

\begin{figure}[!htbp]
\centering
\includegraphics[width=7.8cm]{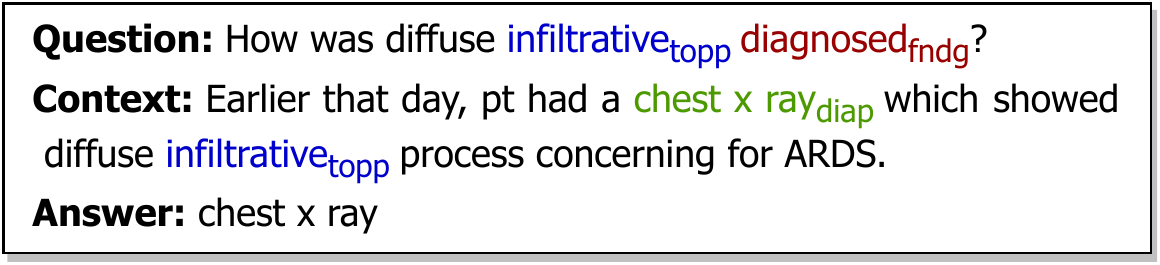}
\caption{An example of a question, context, their extracted entities and expected answer.}
\label{fig:emrqa_ent}
\vspace{-10pt}
\end{figure}

\begin{table}[t]
\scriptsize
\renewcommand{\arraystretch}{1.2}
\centering
\begin{tabular}{llcc}
\toprule
\textbf{Dataset}                      & \textbf{Model}       & \textbf{F1-score} & \textbf{Exact Match} \\ \toprule
\multirow{4}{*}{\textbf{emrQA-s (pl)}} & \textbf{BERT}   & 72.13             & 65.81                \\
                                      & \textbf{cBERT}        & 74.75    (+2.62)         & 67.25 (+1.44)                \\
                                      & \textbf{cERNIE}       & 77.39        (+5.26)     & 70.17 (+4.36)                \\
                                      & \textbf{M-cERNIE} & \textbf{79.87 (+7.74)}             & \textbf{71.86 (+6.05)}                \\ \midrule
                    \textbf{emrQA-s (r)}& \textbf{cBERT} & \textbf{82.34}             & \textbf{74.58}                \\ \midrule
\multirow{4}{*}{\textbf{emrQA-p (pl)}}  & \textbf{BERT}   & 64.19             & 56.30                \\
                                      & \textbf{cBERT}        & 65.45 (+1.26)             & 57.58 (+1.28)                \\
                                      & \textbf{cERNIE}       & 66.15    (+1.96)         & 59.80    (+3.5)            \\
                                      & \textbf{M-cERNIE} & \textbf{67.21 (+3.02)}             & \textbf{61.22 (+4.92)}                \\\midrule
                  \textbf{emrQA-p (r)}& \textbf{cBERT} & \textbf{72.51}             & \textbf{65.14}                \\ \midrule
\multirow{4}{*}{\textbf{MADE-s (pl)}}  & \textbf{BERT}   & 68.45             & 60.73                \\
                                      & \textbf{cBERT}        & 70.19    (+1.74)         & 62.00    (+1.27)            \\
                                      & \textbf{cERNIE}       & 71.51    (+3.06)         & 65.31    (+4.58)            \\
                                      & \textbf{M-cERNIE} & \textbf{73.83 (+5.38)}             & \textbf{67.53 (+6.8)}                \\ \midrule
                  \textbf{MADE-s (r)}& \textbf{cBERT} & \textbf{73.70}             & \textbf{65.54}                \\ \midrule                    
\multirow{4}{*}{\textbf{MADE-p (pl)}}        & \textbf{BERT}   & 63.39             & 57.49                \\
                                      & \textbf{cBERT}        & 64.97    (+1.58)         & 58.94    (+1.45)            \\
                                      & \textbf{cERNIE}       & \textbf{65.71 (+2.32)}             & \textbf{60.55 (+3.06)}                \\
                                      & \textbf{M-cERNIE} & 64.58    (+1.19)         & 59.39 (+1.9)               \\ \midrule
                   \textbf{MADE-p (r)}& \textbf{cBERT} & \textbf{66.89}             & \textbf{61.27}    \\ \bottomrule
\end{tabular}
\caption{F1-score and exact match values for Models on \emph{emrQA} and \emph{MADE}. The `-s' suffix refers to the \emph{sentence} setting and `-p' refers to the \emph{paragraph} setting for the context provided in our reading comprehension style QA task. The `(pl)' refers to the \emph{paraphrase-level} and `(r)' refers to the \emph{random} split as explained in \S~\ref{sec:dataset}. \emph{BERT} refers to $\text{BERT}_{base}$, \emph{cBERT} refers to \emph{clinicalBERT}, \emph{cERNIE} refers to \emph{clinicalERNIE} and \emph{M-cERNIE} refers to the multi-task learning \emph{clinicalERNIE} model. }
\label{tab:exact_match_results}
\vspace{-12pt}
\end{table}
\paragraph{Does clinical entity information improve models' performance?}  
Across all settings, the F1-score of \emph{cERNIE} improves by $\sim2-5\%$ over \emph{BERT} and $\sim0.75-3\%$ over \emph{cBERT}. The exact match performance improved by $\sim3-4.5$ over \emph{BERT} and $1.5-3.25\%$ over \emph{cBERT}. Also, as expected, the performance in \emph{sentence setting (-s)} improved relatively more than it did in \emph{paragraph-setting}. The entity-enriched tokens help in identifying the tokens which are required by the question. For example, in Fig.~\ref{fig:emrqa_ent}, the token `infiltrative' in the question as well as the context get highlighted with the help of the identified entity `topp' (therapeutic or preventive procedure) and then relevant tokens in the context, chest x ray, get highlighted with the relevant entity `diap' (diagnostic procedure). This information aids the model in narrowing down its focus to highlighted diagnostic procedures in the context for answer extraction.


\paragraph{Does logical form information help the model generalize better?} 
In order to answer this question, we compared the performance of our \emph{M-cERNIE} model to \emph{cERNIE} model and observed an improvement of $1.1-2.5\%$ in F1-score and an improvement of $1.4-1.8\%$ in exact match performance. Here as well, the performance improvement is more for \emph{sentence setting (-s)} as compared to the \emph{paragraph setting (-p)}. This helps the model in understanding the information need expressed in the question and helps in narrowing down its focus to certain tokens as the candidate answer. As seen in example~\ref{fig:emrqa_ent}, the logical form helps in understanding that the `dose' of `medication' needs to be extracted from the context where `dose' was already highlighted with the help of the entity embedding of `qnco'. 

Overall, the performance of our proposed model improves the F1-score by $1.2-7.7\%$ and exact-match by $3.1-6.8\%$ over \emph{BERT} model. Thus, embedding clinical entity information with the help of further fine-tuning, entity-enriching and logical form prediction help the model in performing better over the unseen paraphrases by a significant margin. For \emph{emrQA}, the performance of \emph{M-cERNIE} is still below the upper bound performance of the \emph{cBERT} model which is achieved when all the question templates are observed (\emph{emrQA-s/p (r)}) by the model but for \emph{MADE}, in \emph{sentence setting (-s)}, the performance of \emph{M-cERNIE} is even better than the upper bound model performance. For \emph{MADE-p} dataset the performance dropped a little when the LF prediction information is added to the model which might be because \emph{MADE-p} only has $8$ logical forms (Appendix~\ref{appendix:lf_made}) in total, resulting in low variety between the questions. Thus, the auxiliary task did not add much value to the learning of the base model (\emph{cERNIE}) at paragraph level.



\begin{figure}[!htbp]
\centering
\includegraphics[width=7.8cm]{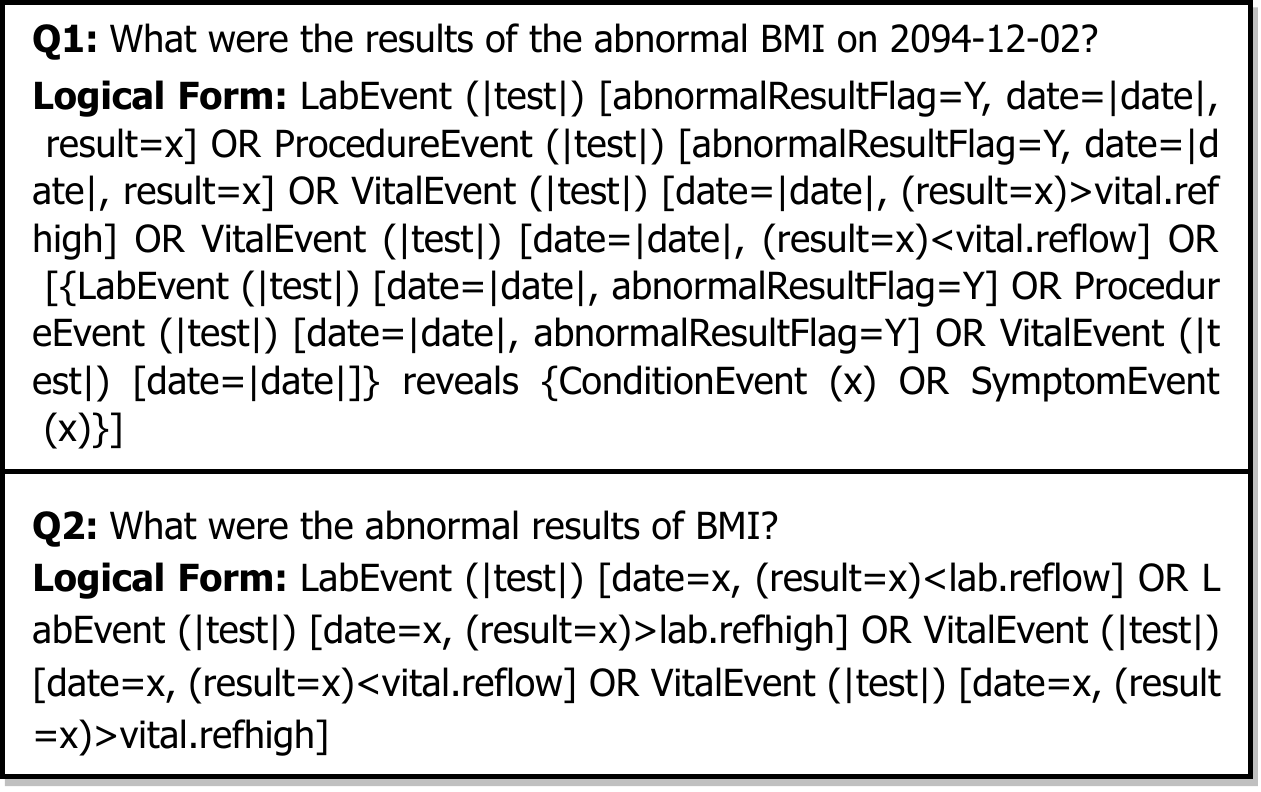}
\caption{Two similar questions with different logical forms (LFs) but overlapping answer conditions.}
\label{fig:lf_sim}
\vspace{-8pt}
\end{figure}

\paragraph{Does the model provide a supporting rationale via logical form (LF) prediction?}
We analyzed the performance of \emph{M-cERNIE} on \emph{MADE-s} and \emph{emrQA-s} datasets for logical form prediction, as we saw most improvement in \emph{sentence setting (-s)}. We calculated macro-weighted precision, recall and F1-score for logical form classification. The model achieved a F1-score of $\sim0.45-0.59$ for both datasets, as shown in Table~\ref{tab:lf_f_score}, \emph{exact} match setting. We analysed the confusion matrix of predicted LF and observed that the model mainly gets confused between the logical forms which convey similar semantic information as shown in Fig.~\ref{fig:lf_sim}.

\begin{figure}[!htbp]
\centering
\includegraphics[width=7.8cm]{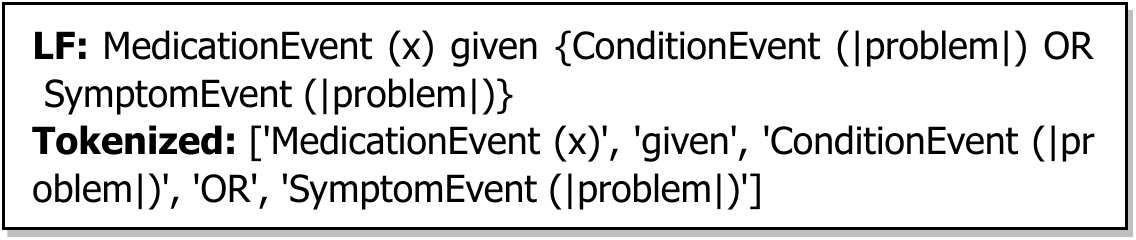}
\caption{Tokenized logical form (LF).}
\label{fig:tokenized_lf}
\vspace{-8pt}
\end{figure}

As we can see in Fig.~\ref{fig:lf_sim} that both logical forms refer to quite similar information, hence, we decided to obtain performance metrics (precision, recall and F1-score) in \emph{relaxed} setting. We designed this \emph{relaxed} setting to create a more realistic setting, where the \emph{tokens} of predicted and actual logical forms are matched rather than the whole logical form. An example of logical form tokenization is shown in Fig.~\ref{fig:tokenized_lf}.

The model achieves a F1-score of $0.92$ for \emph{emrQA-s} and $0.84$ for \emph{MADE-s} in relaxed setting (Table~\ref{tab:lf_f_score}). This suggests that the model can efficiently identify important semantic information from the question, which is critical for efficient QA. During inference, the \emph{M-cERNIE} models yield a rationale regarding a new test question (unseen paraphrase) by predicting the logical form of the question as an auxiliary task. For ex, the LF in Fig.~\ref{fig:emrqa_qa} provides a rationale that any \textit{lab} or \textit{procedure} event related to the \textit{condition} event needs to be extracted from the EMR for diagnosis.
\begin{table}[!htbp]
\centering
\scriptsize
\renewcommand{\arraystretch}{1.1}
\begin{tabular}{l|l|ccc}
\toprule
\textbf{Setting}                                 & \textbf{Dataset} & \textbf{Precision} & \textbf{Recall} & \textbf{F1-score} \\ \toprule
\multirow{2}{*}{\textbf{Exact}}      & \textbf{emrQA}   & 0.65              & 0.61          & 0.59            \\
                                                 & \textbf{MADE}    & 0.47              & 0.52           & 0.45             \\ \midrule
\multirow{2}{*}{\textbf{Relaxed}} & \textbf{emrQA}   & 0.93              & 0.91           & 0.92             \\
                                                 & \textbf{MADE}    & 0.83              & 0.85           & 0.84            \\ \bottomrule
\end{tabular}
\caption{Precision, Recall and F1-score for logical form prediction.}
\label{tab:lf_f_score}
\vspace{-12pt}
\end{table}



\paragraph{Can logical form information be induced in multi-class QA tasks as well?}
To answer this question, we performed another experiment where the model has to classify the evidence sentences from the non-evidence sentences making it a two-class classification task. The model would be provided a tuple of question and a sentence and it has to predict whether the sentence is evidence or not? The final loss of the model ($\mathcal{L}_{model}$) changes to:
\begin{equation}
    \mathcal{L}_{model} = \omega \mathcal{L}_{lf} + (1-\omega) \mathcal{L}_{evidence}
\end{equation}
where $\omega$ is the weightage given to the loss of auxillary task ($\mathcal{L}_{lf}$), logical form prediction. $\mathcal{L}_{evidence}$ is loss for evidence classification and $\mathcal{L}_{model}$ is the final loss for our proposed model. We conducted our experiments on \emph{emrQA} dataset as evidence sentences were provided in it. In the multi-class setting, the \verb+[CLS]+ token representation would be used for evidence classification as well as logical form prediction.

\begin{table}[!htbp]
\centering
\scriptsize
\renewcommand{\arraystretch}{1.1}
\begin{tabular}{l|l|ccc}
\toprule
\textbf{Dataset}                          & \textbf{Model}       & \textbf{Precision} & \textbf{Recall} & \textbf{F1-score} \\\toprule
\multirow{3}{*}{\textbf{emrQA}} & \textbf{cBERT}        & 0.67              & 0.99           & 0.76            \\
                                          & \textbf{cERNIE}       & 0.69              & 0.98           & 0.78 (+0.02)            \\
                                          & \textbf{M-cERNIE}  &  0.73            &     0.99    &   \textbf{0.82 (+0.06)} \\\midrule
\end{tabular}
\caption{Macro-weighted precision, recall and F1-score of Proposed Models on Test Dataset (Multi-choice QA). For the model names, c: clinical; M: multitask.}
\label{tab:prf_results}
\vspace{-13pt}
\end{table}

The multi-task entity enriched model (\emph{M-cERNIE}) achieved an absolute improvement of $6\%$ over \emph{cBERT} and $4\%$ over \emph{cERNIE}. This suggests that the inductive bias introduced via LF prediction does help in improving the overall performance of the model for multi-class QA as well. 


\section{Related Work}
\label{sec:rel}


In the general domain, BERT-based models are on the top of different leader boards across various tasks, including QA tasks~\cite{rajpurkar2018know,rajpurkar2016squad}. The authors of \cite{nogueira2019passage} applied BERT to the MS-MARCO passage retrieval QA task and observed improvement over state of the art results. \cite{nogueira2019document} further extended the work by combining BERT with re-ranking of predictions for queries that will be issued for each document. However, BERT-based models have not been adapted to answering physician questions on EMRs.

In case of domain-specific QA, logical forms or semantic parse are typically used to integrate the domain knowledge associated with a KB-based (knowledge base) structured QA datasets, where a model is learnt for mapping a natural language question to a LF. GeoQuery~\cite{zelle1996learning}, and ATIS~\cite{dahl1994expanding}, are the oldest known manually generated question-LF annotations on closed-domain databases. QALD~\cite{lopez2013evaluating}, FREE 917~\cite{cai2013large}, SIMPLEQuestions~\cite{bordes2015large} contain hundreds of hand-crafted questions and their corresponding database queries. Prior work has also used LFs as a way to generate questions via crowd-sourcing~\cite{wang2015building}. WEBQuestions~\cite{berant2013semantic} contains thousands of questions from Google search where the LFs are learned as latent representations in helping answer questions from Freebase. Prior work has not investigated the utility of logical forms in unstructured QA, especially as a means to generalize the QA model across different paraphrases of a question.

There have been efforts on using multi-task learning for efficient question answering, such as the authors of \cite{mccann2018natural} tried to learn multiple tasks together resulting in an overall boost in the performance of the model on SQuAD \cite{rajpurkar2016squad}. Similarly, the authors of \cite{lu201912} also utilised the information across different tasks which lie at the intersection of vision and natural language processing to improve the performance of their model across all tasks. The authors of \cite{rawat2019naranjo} utilised weak supervision to the model while predicting the answer but not much work has been done to incorporate the logical form of the question for unstructured question answering in a multi-task setting. Hence, we decided to explore this direction and incorporate the structured semantic information of the questions for extractive question answering.


\section{Conclusion}
The proposed entity-enriched QA models trained with an auxiliary task improve over the state-of-the-art models by about $3-6\%$ across the large-scale clinical QA dataset, emrQA \cite{pampari2018emrqa} (as well as MADE \cite{jagannatha2019overview}). We also show that multitask learning for logical forms along with the answer results in better generalizing over unseen paraphrases for EMR QA. The predicted  logical forms also serve as an accompanying justification to the answer and help in adding credibility to the predicted answer for the physician.

\section*{Acknowledgement}
This work is supported by MIT-IBM Watson AI Lab, Cambridge, MA USA.

\bibliography{acl2020.bib}
\bibliographystyle{acl_natbib}

\clearpage
\appendix

\section{Model Hyper-parameters}
\label{appendix:hyper}
Most of the hyper-parameters across our models remained same: learning rate: $2e-5$, weight decay: $1e-5$, warm-up proportion: $10\%$ and hidden dropout probability: $0.1$. The parameters that varied across models for different datasets are enumerated in the Table~\ref{tab:hyper}. The hyper-parametsrs provided in Table~\ref{tab:hyper} are for all models in a particular dataset. This also suggests that even after adding an auxiliary task, the proposed model doesn't need a lot of hyper-parameter tuning. 
\begin{table}[!htbp]
\small
\centering
\begin{tabular}{lcc}
\toprule
\textbf{Dataset}          & \textbf{\begin{tabular}[c]{@{}c@{}}Entity Embedding \\ Dim\end{tabular}} & \textbf{\begin{tabular}[c]{@{}c@{}}Auxiliary Task \\ Wt.\end{tabular}} \\ \toprule
\textbf{emrQA-rel} & 100                                                                      & 0.3                                                                    \\
\textbf{BoolQ}     & 90                                                                       & 0.3                                                                    \\
\textbf{emrQA}     & 100                                                                      & 0.3                                                                    \\
\textbf{MADE}      & 80                                                                       & 0.2                                                                   \\ \bottomrule
\end{tabular}
\caption{Hyper-parameter values across different datasets.}
\label{tab:hyper}
\end{table}

\section{Logical forms (LFs) for MADE dataset}
\label{appendix:lf_made}
1. MedicationEvent ($|$medication$|$) [sig=x]\\
2. MedicationEvent ($|$medication$|$) causes {ConditionEvent (x) OR SymptomEvent (x)} \\
3. MedicationEvent ($|$medication$|$) given {ConditionEvent (x) OR SymptomEvent (x)}\\
4. [ProcedureEvent ($|$treatment$|$) given/conducted {ConditionEvent (x) OR SymptomEvent (x)}] OR [MedicationEvent ($|$treatment$|$) given {ConditionEvent (x) OR SymptomEvent (x)}]\\
5. {MedicationEvent (x) CheckIfNull ([enddate]) OR MedicationEvent (x) [enddate$>$currentDate] OR ProcedureEvent (x) [date=x]} given {ConditionEvent ($|$problem$|$) OR SymptomEvent ($|$problem$|$)}\\
6. {MedicationEvent (x) CheckIfNull ([enddate]) OR MedicationEvent (x) [enddate$>$currentDate]} given {ConditionEvent ($|$problem$|$) OR SymptomEvent ($|$problem$|$)}\\
7. {MedicationEvent ($|$treatment$|$) OR ProcedureEvent ($|$treatment$|$)} given {ConditionEvent (x) OR SymptomEvent (x)}\\
8. {MedicationEvent ($|$treatment$|$) OR ProcedureEvent ($|$treatment$|$)} improves/worsens/causes {ConditionEvent (x) OR SymptomEvent (x)}

\section{Selected entities from MetaMap}
\label{appendix:sem_types}
The list of selected semantic types in the form of entities and their brief descriptors are provided in Table~\ref{tab:sem_defs}.
\begin{table}[]
\small
\begin{tabular}{ll}
\toprule
Semantic Type & Description\\ \midrule
acab & Acquired Abnormality                 \\ 
aggp & Age Group                            \\
anab & Anatomical Abnormality               \\
anst & Anatomical Structure                 \\
bpoc & Body Part, Organ, or Organ Component \\
cgab & Congenital Abnormality               \\
clnd & Clinical Drug                        \\
diap & Diagnostic Procedure                 \\
emod & Experimental Model of Disease        \\
evnt & Event                                \\
fndg & Finding                              \\
inpo & Injury or Poisoning                  \\
lbpr & Laboratory Procedure                 \\
lbtr & Laboratory or Test Result            \\
phob & Physical Object                      \\
qnco & Quantitative Concept                 \\
sbst & Substance                            \\
sosy & Sign or Symptom                      \\
topp & Therapeutic or Preventive Procedure  \\ \bottomrule
\end{tabular}
\caption{Selected semantic types as per MetaMap and their brief descriptions.}
\label{tab:sem_defs}
\end{table}

~~~~~~~~~~~~~~~~~~~~~~~~~~~~~~~~~~~~~~~~~


\end{document}